# The many faces of deep learning

**Raul Vicente**[1]

[1]Computational Neuroscience Lab, Institute of Computer Science, University of Tartu, Tartu, Estonia

**\* Correspondence:**
Raul Vicente
raulvicente@gmail.com

**Keywords: deep learning, artificial intelligence, mathematics, physics, neuroscience.**

**Abstract**

Deep learning has sparked a network of mutual interactions between different disciplines and AI. Naturally, each discipline focuses and interprets the workings of deep learning in different ways. This diversity of perspectives on deep learning, from neuroscience to statistical physics, is a rich source of inspiration that fuels novel developments in the theory and applications of machine learning. In this perspective, we collect and synthesize different intuitions scattered across several communities as for how deep learning works. In particular, we will briefly discuss the different perspectives that disciplines across mathematics, physics, computation, and neuroscience take on how deep learning does its tricks. Our discussion on each perspective is necessarily shallow due to the multiple views that had to be covered. The deepness in this case should come from putting all these faces of deep learning together in the reader's mind, so that one can look at the same problem from different angles.

## 1. Introduction

Why deep learning works so well? The typical answer to this question depends on to which community is addressed. Mathematicians will rightly contend that we do not really know. Apart from theorems concerning the expressibility of neural networks (e.g. they are universal approximators of continuous functions given some mild conditions), there are hardly any mathematical guarantees of their ability to learn when deployed on real-world problems. This is not surprising given that we cannot have a precise mathematical characterization of how real data looks like. That would probably involve modeling the data-generating process or in other words, approximate how the laws of physics conspire to produce complex patterns in the data that we collect about ourselves and our environment. How one would capture in a few manageable equations the precise distribution of pixels and object categories in an arbitrary set of natural or medical images? Without such a mathematical characterization of the data, or assuming some wildly rough approximation, it is difficult to know the exact landscape against which optimization methods have to navigate, and hence to provide sharp learning and generalization guarantees.

This does not mean that useful intuitions and insights have not been developed in different communities. After all, practitioners of deep learning do find successful models for real large-scale problems, and while it is usually claimed that training neural networks is still an art more than a science, it is not a blind shot either (see Schmidhuber, 2015 for a historically detailed review). Even without mathematical guarantees, many mathematical and technical concepts in computer science, optimization, statistics, physics, and neuroscience have inspired and continue being a guide to understand how deep learning achieves its results.

In this work, we wish to collect and synthesize different intuitions scattered across several fields as for how deep learning works. The main goal of discussing these interpretations from different areas is akin as to why several proofs of the same theorem are useful: they give different perspectives to the same object one wishes to understand better.

Given the diversity of current architectures and algorithms, we will focus our attention on the simplest setting and consider a feedforward neural network trained by supervised learning (e.g. data consisting of a set of exemplary input-output pairs). Many architectures and types of learning are variations building upon this fundamental configuration. For example, recurrent neural networks can be understood as feedforward architectures unfolding over time, while most of the cleverness in several algorithms of unsupervised and reinforcement learning goes into how to efficiently convert the problem into one of supervised nature.

Importantly, in our presentation we will focus on what makes deep learning special from other machine learning methods. Hence, out of the four components of any machine learning algorithm (data, model, cost function and optimization method), we will mostly deal with the model component since it is in this aspect where deep learning makes particular choices compared to other algorithms (Goodfellow et al., 2016). The model or hypothesis space considered in deep learning consist of the family of functions obtained by composing a finite number of non-linear functions with adjustable parameters or weights.

Before we proceed to describe different intuitions about what such a composition of adaptive functions does to your data, we will need to set some background and terminology that will be useful for the rest of our discussion. Unless stated otherwise, we will consider a typical dataset with $k$ samples and $n$ numerical features that can be described as a $k$ x $n$ design matrix or more geometrically as a set of $k$ points in an $n$ dimensional Euclidean space. In the supervised setting we will consider either a classification problem for which there is a discrete number $c$ of labels or classes (which can be thought of as different colors of the sample points), or a regression problem with real numerical values as targets. The goal of the training is then to produce a function, mapping inputs to outputs, such that a given cost function or error is minimized. In the probabilistic setting the output of the learned function is used as parameter such as, for example, the mean in a conditional probability model describing the probability of a range of outputs given a certain input.

## 2.  Perspectives

Different disciplines have naturally developed distinct interests and points of view in regard to the theory and application of deep learning. Here, we wish to expose the varied views on deep learning phenomena by discussing a few illustrative examples from each of the considered perspectives.

### 2.1.  Topological perspective

While random numbers are a precious computational resource, most data we generate or care about to measure contains certain structure. For example, the intensities of neighboring pixels in a picture are not independent, they usually exhibit similar values or low-frequency patterns (textures) that extend over large regions and objects across the image. These correlations imply that out of the set of all images that one could possibly imagine, we only have to deal with a fraction of structured images which exhibit numerous and strong correlations among its features.

More generally, when a dataset is considered as a set of points these will typically form a shaped cloud or manifold which only occupies a subspace of the possible volume of the *n* dimensional space spanned by the different feature dimensions. By manifold here we do not mean the usual definition in topology (a surface that locally looks like Euclidean space near each point) but rather a slab of connected points that is relatively thin in many directions (Goodfellow et al., 2016). There can even be several of such slabs. The important point is that this cloud can be of a very complicated and convoluted shape, and its division in regions where points belong to the same class (or colouring) can also be complicated.

From a topological point of view one can ask what transformations does this cloud undergo when represented by successive layers of a neural network. The weighted sum of inputs to a layer (affine transformation) simply stretches the cloud in some directions and compresses it in others. The additive bias terms have the effect of shifting or translating the resulting cloud. However, simply composing these linear transformations will not produce too useful transformations to disentangle a complicated manifold or its colouring. Here is where the non-linearity of the activation of neurons or units becomes essential. A non-linear activation function maps a range of inputs into a range of outputs with certain "distortion". In non-linear functions proportionality cannot be maintained for the whole range of inputs, so, for example, doubling the input will not always result in doubling the output that is produced by the function. This means that the shape of our cloud of points will be bent in some directions due to the non-linearity of the activation functions. An extreme case of such distortions occurs, for example, when using the absolute value function as the activation function. In that case, the V-shaped function makes that pairs of input values (mirror points at each side of the bottom of the V) are transformed into the same output value. So, in effect, this transformation produces a perfect folding of the cloud along some axis because several points in the input are projected to the same point at the output, as when one folds a piece of paper. As we proceed with the forward pass through the layers of the network and repeat this operation, one is simply composing one folding over previous ones (Montufar et al, 2014). The training of the network thus becomes interpreted as the learning of along which directions to fold our successive representations of the cloud to disentangle the colouring of its points into regions that are linearly separable (Montufar et al, 2014). Similar interpretations hold for more usual activation functions such as ReLU or sigmoid. In a loose sense, the training of a network becomes an exercise in high-dimensional origami on some elastic material!

The former interpretation is useful to understand how the chaining of simple transformations can take a difficult classification problem in the original space to a much simpler one in a new distorted space. Moreover, the composition of elementary transformations, each building on top of the previous one, allows to unfold complex dependencies across distant regions of the cloud of data while using a reduced number of training samples. However, note that since the composition of continuous functions is just another continuous function, no matter how much we play with the weights, the transformation from inputs to outputs will send neighboring points to neighboring points, and thus cutting a cloud in two or more pieces is not an operation allowed by continuous activation functions. Everything must be obtained from stretching, compressing and bending. Given the visual intuition and understanding provided by topological and geometrical thinking, it is interesting to entertain what are the topological effects occurring through an autoencoder's bottleneck, or in the transformations produced by skip connections and residual layers found in more sophisticated architectures.

So far we have considered the problem of discriminating or classifying points into different classes. A more ambitious challenge is that of building a generative model of the data distribution where new samples and other manipulations beyond the discrimination of classes can be obtained. A popular architecture to produce generative models of real data is the Generative Adversarial Network or GAN (Goodfellow et al, 2014). Its training proceeds as a forger-police zero-sum game in which one network (the generator) tries to produce samples that look like the training dataset, while another network (the discriminator) tries to detect the counterfeit. Regardless of the specifics of its training the generator network can be thought as being fed a random sample in some latent space and producing a real-data looking sample. Similarly as in the previous case, this very complex transformation can also be understood in geometric terms. In this case, the input consists of a cloud with the shape of some prior distribution such as Gaussian noise. The task of the successive layers of the generator network is to convert this cloud into the target manifold of the training data. It is perhaps not surprising then, that the same architectures capable of disentangling a manifold into linearly separable regions, can also be trained to produce the stretching and folding necessary to forge an undifferentiated cloud into convoluted manifolds spanned by real data. The intended effect of the network is thus akin to a pizza maker who starting from a ball of dough stretches and folds it until producing some desired shape. One must note however that the training is not without difficulties and much effort has been invested in assisting the network to avoid mode collapse (Arjovsky et al, 2017), the undesired fact that the transformation might end up targeting a particular region or only one of the possible manifolds.

As a final case, it might be also elucidating to search for a topological interpretation of the phenomena of adversarial examples. Adversarial examples refer to certain small and almost unnoticeable (at least for humans) variations in the input which drastically change the output of a classifier, and they pose a significant challenge to the practical application of neural networks as well as other machine learning models (Yuan et al, 2019; Athalye et al, 2018). Given that the transformations a neural network can perform are continuous, it still must be the case that neighborhoods in the input space are mapped into neighborhoods in the successive representations. What is then responsible for the existence of adversarial examples? It is important to note that the notion of continuity actually refers to sufficiently small neighborhoods. While the changes in adversarial examples might look unnoticeable to the human eye, they do not violate the continuity property since the perturbations are not arbitrarily small. Indeed, it has been proposed that the original input and its adversarial perturbation can diverge into regions labeled with distinct classes due to the almost-linearity of transformations realized by the layers of the network (Athalye et al, 2018). That occurs for example when most ReLU units are activated within their linear range. The repeated application of almost linear or affine maps can amplify small differences in certain directions of the original input space so that the original point and its targeted perturbation end up in separated regions when represented by the final layers (Goodfellow et al, 2014b). Interestingly, non-linear transformations could in principle also contribute to adversarial examples if the training is not fully optimal. The reason is that the same recurrent stretching and folding process that is intended to produce linearly separable classes is also a hallmark of the non-linear mixing that gives rise to the sensitivity to initial conditions found in chaotic dynamical systems.

Adversarial examples are still poorly understood and the arms race between adversarial attacks and the fixes to protect the models from such attacks keeps escalating. In any case, topological and geometrical intuitions are a powerful guide for understanding how deep learning works as well as how it fails.

## 2.2. Metric Perspective

In many cases the encoding of input lacks any information about the similarity or distance between different samples. For example, symbolic representations of words do not have any resemblance to the objects or meanings they stand for. A potato and a bathtub are just as similar as horse and zebra in what concerns their symbolic representations.

In particular, one-hot encodings represent different items by the position of a single "1" within a binary vector with "0"'s in the rest of its entries. In essence, this representation just uses a mask with one "1" to single out distinct categorical variables, and thus the length of this representation is as large as the number of different types of items, which can be dramatically large in many applications. In our point cloud view each different data point sits on a different orthogonal axis and is just as far away from any other point. This representation is hardly useful to tackle questions about interpolation or generalization precisely because the lack of a meaningful notion of similarity or distance between these points or even of the space spanned by them.

Neural network embeddings address this problem by learning a representation that encodes each sample as a relatively short vector with continuous values (see Camacho-Collados, J. and Pilehvar, M.T., 2018 for a review). A neural network embedding can be obtained by using a short hidden layer to produce a continuous representation on the way of solving some supervised task. In a classical example, a simple network with a linear hidden layer is given one-hot encodings of a sequence of words and is trained to output the one-hot encoding of the next word (Mikolov et al, 2013). The one-hot encoding of the raw input multiplied by the weight matrix corresponds to selecting a row of the matrix. Such weights matrix contains real continuous values, and after training, its rows are the representation for each different type of input. In this case of word embeddings, the prediction task forces that words that appear in similar contexts to get multiplied by similar weights, and therefore represented by similar activations in the hidden layer. This way the inputs are represented by short and dense vectors which relative coordinates capture useful information. Indeed, the learned representations have been observed to acquire relevant semantic structure (Mikolov et al, 2013). For example, in word embeddings the words of similar meaning are mapped to nearby vectors, and directions in the new space can correspond to particular relations between words. Similar phenomena occur for more sophisticated embeddings of full sentences and paragraphs.

The gain here is obtained not so much from the depth (the architecture in the example above is actually shallow since it used one linear hidden layer) as from the short and distributed nature of the learned representation. This distributed representation is just a usual vector which lives or is embedded into a Euclidean space of the same dimension as the length of the hidden layer. Hence, the new cloud of points or vectors is naturally equipped with the structures to measure distances and angles of the Euclidean space in which it is embedded. In other words, thanks to the training of the embedding a meaningful metric or similarity function between points is now available. This implies the possibility to exploit notions such as neighborhood, an essential concept to support similarity and interpolation reasoning, as well as the general metric properties of Euclidean or manifold spaces. Moreover, in addition to the extrinsic notions of distance (inherited from the ambient space), it is also possible to explore whether intrinsic metric properties defined on the data manifold, such as for example geodesic or diffusion distances, capture additional structure of the data.

Importantly, what previously was impossible with one-hot encodings can become possible with embedded representations. Equipped with distributed representations one can describe the position of a novel point in coordinates relative to the positions of other points. This relational aspect is key for generalization and even one-shot transfer learning since training points can serve as anchor points for describing novel inputs and inferring unseen correspondences between different domains (Goodfellow et al, 2016). Moreover, embeddings can also be applied to create join representations of inputs belonging to different modalities, such as an image and its accompanying text annotation, which otherwise are difficult to compare or align. Thus, the metric and similarity notions induced by embedded representations might be one of the fundamental ways to understand how humans and machines support analogy reasoning and even high-level concepts.

### 2.3. Information Perspective

In the probabilistic setting supervised learning can be viewed as the learning of a conditional probability model p(Y|X) that predicts the likelihood of different targets (Y=y) given an input (X=x). Thus, inputs and outputs in our training data can be considered as samples obtained from a pair of random variables, X and Y, which the model has to relate. This is possible only if X contains information about Y, that is, if uncertainty about Y is reduced when observing X, a measure which is quantified using mutual information (Cover, T.M. and Thomas, J.A., 2012). An interesting question is how the successive layers in a neural network affect the amount of information retained about the original input (Tishby, N. and Zaslavsky, N., 2015).

In standard feedforward networks, each layer applies a transformation to the input received from the previous layer. Thus, the set of layers, from input to output, form a Markov chain in which the activation values of each layer only depend on the previous ones via the immediately preceding layer (Tishby, N. and Zaslavsky, N., 2015). In particular, whenever a layer involves a non-linear activation function that is not invertible (two or more inputs produce the same activation, such as in the flat part of a ReLU) or there is a reduction in the number of units compared to the previous layer (more generally when the weight matrix has no inverse), the result is that the transformation is non-invertible. The non-invertibility implies that activations of the previous layer cannot be recovered from the set of activations of the present layer. Indeed, most interesting transformations or mathematical functions, including the usual sum of two numbers, are non-invertible. For example, in the case of the sum, given the result of the operation (7) one cannot unequivocally determine which were the specific inputs since several input combinations (3+4=2+5) are compatible with the same output result. For the purpose of the computation such set of inputs are indistinguishable, or in other words, the computation output is invariant against changes within such set of inputs.

From an information point of view each layer is literally throwing away or discarding information that was present in the input. Thus, the representations by intermediate layers cannot increase the amount of information about the target beyond that already contained in the input. While at first it might sound counterintuitive that learning is based upon throwing away or forgetting information, indeed discarding information is an essential feature of computation or information processing (Doya et al, 2007). The key is to selectively retain and make explicit the information that is relevant for predicting the target while throwing the rest away. This is indeed the hallmark of the invariance necessary to, for example, being able to recognize the same person regardless of her distance, face orientation, hair style, illumination conditions, etc. In a sense, any interesting computation consists of carving out the relevant information from a sea of irrelevant one. The successive transformations carried out by a neural network during its training process precisely aim to find the combination of features that explicits out information (linearly extractable by the output layer) that is best correlated with the desired target. In the process, dimensions or features that contain noisy or irrelevant information with respect to the target classes have to be folded or projected out so their value does not affect the final representation, which is the geometric analog of throwing such information to the trash bin.

Moreover, it has been proposed that the more compressed the learned representations get, while still retaining information about the target, the better are their generalization properties (Tishby, N. and Zaslavsky, N., 2015). Thus, it might be possible to interpret that the optimal training of a neural network aims to produce representations that are close to a minimally sufficient statistic of the input with respect to the target random variable. This can be considered as some reflection of Occam principle by which, other things equal, simpler or shorter rules tend to avoid overfitting and generalize better (Calude, C.S. and Chaitin, G.J., 2007).

More generally, it might be interesting to assess how the information contained in different parts or components of the input interact to explicit out relevant information at the output layer. In particular, one part of the input can contain information about the target that is unique, redundant, or synergistic with respect information from another part of the input (Bertschinger et al, 2014). How the training of neural networks acts on these channels of information or whether they can be exploited for better performance is currently unknown. In any case, we find that the use of information theoretic concepts and functionals is an important tool to understand and guide the training of deep learning architectures.

### 2.4. Causal Perspective

Recently, deep learning has been criticized as being nothing else than fancy curve fitting. As provoking as it sounds, the critics do not wish to diminish the enormous practical value of deep learning in large-scale problems. Rather, they point out that current deep learning models work at the level of associations or correlations between inputs and outputs, describe the limitations inherent to operating at such a level, and propose causal models as the necessary level to overcome such limitations (Pearl, 2019).

Indeed, in the supervised setting deep learning and other statistical models parametrize a conditional probability $p(Y|X)$ to predict the likelihood of a set of outcomes after observing a particular input. While the model can be more or less sophisticated, the learned conditional probability amounts to an interpolation by fitting parameters based on observed statistical associations between X and Y. Only relying on these passive observations, critics argue, is difficult to empower a learning system with many of the essential characteristics of human-like intelligence (Pearl, 2019). For example, we humans are relatively good at adapting to domain or task shifts (transfer learning) even if only a few samples are available in the novel domain, a feat that current learning systems can hardly achieve despite numerous efforts. We are also able to reason why we make a certain prediction and how it would change if conditions would have varied (explainability). More generally, these and other tasks are thought to be solved by humans by learning and employing causal models (Lake et al, 2017).

Efficient learning systems displaying these desired properties will likely require to also go beyond the level of passive observations and endowing them with abilities to discover and model causal relations of the data generating process or environment. That is, causal reasoning, the ability of identifying a relationship of cause and effect, has been put forward as a fundamental challenge towards human-like AI (Guo et al, 2018; Pearl, J. and Mackenzie, D., 2018). Discovery of causal relations requires actors to either be able to intervene on the environment or have a model of the world accurate enough to answer questions of the kind "what would have happened if instead of doing A I would have done B?", i.e. to imagine an alternative world where something different had happened. These causal and counterfactual questions are a hallmark of human cognition and the building blocks to generalize across different domains with related causal structures or form explainable decisions and plans.

Do deep learning systems learn anything related to a causal model? In machine learning, the most natural setup for studying these questions is that of reinforcement learning (RL) (for an introduction to RL see Sutton, R.S. and Barto, A.G., 2018). In reinforcement learning an agent repeatedly interacts with an environment (real-world or simulated) by producing an action and receiving a new partial observation of the environment state and possibly a reward. In this general framework the agent is trained to learn a policy, that is a mapping between observations to actions, in order to maximize its cumulative reward. Thus, RL agents can learn to model relations between their actions and state transitions and rewards in the environment. In particular, deep nets in RL are used as function approximators to mediate the learning, storage and interpolation of input representations to predict the value of states and actions or explicitly represent a policy (Arulkumaran et al, 2017).

Since an agent in RL can intervene on the environment with its own actions, build a world model (model-based RL), and in some algorithms even decouple the evaluation of its policy from its own actions (off-policy methods), RL models have the potential to reach certain, although perhaps not the highest and more formal, levels of causal inference. In most cases the related causal problem of credit assignment in RL, that is how actions and states in the past contribute to the current reward, is only dealt by the passive diffusion of value from the reward to events in the recent past. Thus once a reward is obtained by the agent, states and actions visited in the most recent past change their value according to the magnitude and their temporal distance to the reward event. States and actions that repeatedly participate in recent trajectories leading to the obtention of reward will start to be singled out as valuable. Without being augmented by any strict causal model nor equipped with any other sort of information or bias, the assignation of credit or blame to different states and actions is a slow and passive process.

From a human perspective, one often finds that an RL agent seems to exploit spurious correlations or shortcuts during learning that differ from the causal links that humans search for and recognize. This is partly due to the large amount of inductive biases that humans assume from previous experiences that are not accessible to agents trained only for playing one particular environment or video game (Zador, 2019). From an agent's point of view, that is the whole world that it has ever existed and any cue with predictive value for that environment is as good as any other, regardless of our human causal interpretations. However, it seems reasonable that picking up "unintended" correlations comes with a price in more realistic and changing environments. In these cases, true causal links describing mechanistic relations are more likely to remain invariant, and therefore, useful for generalizing across tasks or environments. For example, cardiac muscle compression (A) leads to both pumping of blood (B) and heartbeat sound (C). The latter correlates with cardiac function and is commonly used for monitoring its state. But in case of sudden heart malfunctioning we will certain like to act on the mechanisms of cardiac muscle compression rather than on the soundtrack.

Thus, distinguishing causal patterns such as common drive effects (A implies both B and C) and direct (A implies C) vs indirect interactions (A implies B which in turn implies C) have implications for the efficiency of learning and generalization to changing or diverse conditions. The fact that humans are exposed to many different tasks and environments probably adds pressure for us to discover and intervene on causal links since they are likely to be more consistent and useful across tasks than mere correlations. However, in simple and stationary environments, in which most RL algorithms are trained, the pressure and tools for agents to discover and transfer causal relations remains limited. An interesting avenue to add the necessary pressure and tools can be the use of intrinsic rewards for causal discovery and the incorporation of heuristics used by humans and other animals in their modeling of causal relations (Oudeyer, P. Y. and Kaplan, F., 2009; Johnson, S.G. and Keil, F.C., 2014). While imperfect and not formally as strict as "do calculus" or structural causal modeling, it seems that these ingredients have given our brains a boost in their handling of causal relations in natural environments.

In summary, regardless of how refined the state representations obtained by deep learning architectures can be, if unassisted by particular pressures and tools they might not be enough for agents to jump to higher levels of the causal ladder (Pearl, J. and Mackenzie, D., 2018). Endowing deep learning agents with pressures and tools to actively interrogate and build causal models of the dynamics of the environment and of other agents is one of the most significant challenges ahead. An immediate reward will be a better understanding of how transfer learning and generalization depends on how well agents internalize the causal structure of the world around them.

### 2.5.   Physics Perspective

One way to understand an artificial neural network is as a large collection of simple interacting units which cooperate to represent a function or probability distribution. A large ensemble of interacting units, such as molecules in a gas, is also the natural setting for statistical physics, and ideas from this community continue being influential on the theory and numerical algorithms of neural networks (Sompolinsky, 1988; Mézard, M. and Mora T., 2009; Advani et al, 2013, Goodfellow et al, 2016).

In particular, statistical physics aims to relate the collective behavior of a system with the gross statistical properties of its individual parts and their interactions. To do so, physicists define a scalar function called Hamiltonian or energy E which encodes how the parts interact with each other (often guided by some symmetry considerations) and maps every possible configuration of the parts to some real number. Importantly, the probability of finding the system in a configuration decreases exponentially with its energy level according to a Boltzmann distribution. In statistical physics the weights or couplings between the elementary units are usually fixed and one is interested in averaging or sampling from the associated Boltzmann probability distribution in order to derive the macroscopic properties of interest.

An important family of models in machine learning receives the name of energy-based models precisely because they use the Boltzmann distribution of some energy function to parametrize a probability distribution (LeCun et al, 2006). These models include Boltzmann machines which use a family of energy functions known as Ising models which originated in the study of magnetism as a collective property of a large collection of atoms (Ackley et al, 1985). Neurons in these models, similar to magnetic spins, can be understood as binary threshold units driven by external input as well as by the weighted output of similar units with which they are coupled.

While in statistical physics the weights are usually fixed and obtaining averages or samples from the Boltzmann distribution is the main goal, in machine learning the interest is in learning and inference. Thus, during learning of energy-based models the focus is on finding the weights between units so that the resulting Boltzmann distribution maximizes the likelihood of the training dataset. This is equivalent to search in the space of allowed energy functions for the weights that minimize the energy of the training data. Every term in the resulting energy function works as a soft constraint between variables by favouring some correlations and discouraging others in order to model the dependencies existing between variables in the data (Hinton, 2002). During inference, after the weights have been fixed, all observable variables or inputs are clamped and the rest of variables sampled from the Boltzmann distribution (favouring states that minimize the energy function). Reading out these inferred values can be used to make predictions given the observed data, or recall patterns that were stored as minimum-energy states.

Restricted Boltzmann Machines (RBM) are a type of Boltzmann machine with a bipartite connectivity between visible and hidden nodes which makes them efficient to train. Stacking several RBMs and training them in a layer-wise manner was one of the first tractable procedures to learn a hierarchy of representations (Salakhutdinov, R. and Hinton, G., 2009). This was indeed one of the essential steps in the pre-training of some of the early successful deep learning models. Moreover, the connection between RBMs and statistical physics has been strengthened by the realization that in such models different layers can be put in correspondence with the iterations of certain renormalization group mapping (Mehta, P. and Schwab, D.J., 2014; Koch-Janusz, M. and Ringel, Z., 2018). This later procedure describes the behavior of a system when it is viewed at progressively larger scales and offers yet another interpretation of how a stack of RBM layers processes information.

Other relevant deep learning networks, in this case using ReLU units, have also been mapped to spin glass models studied in statistical physics of disordered systems (Choromanska et al, 2015). Such mappings allow the transfer of many theoretical and numerical results from a well studied physical system to recent deep learning models. In particular, these results have provided a better understanding of the energy landscapes occurring in deep learning models and why their local minima are typically as good as finding the intractable absolute minimum (Dauphin et al, 2014; Choromanska et al, 2015).

Another example of fruitful interaction between physics and machine learning concerns the role of noise. The effects of the injection of noise or intrinsic random fluctuations has been a subject of study in statistical physics and dynamical systems. Interestingly, in many conditions the injection of certain amount of noise is observed to have constructive role rather than detrimental effects (Simonotto et al, 1997; Gammaitoni et al, 1998; Pikovsky, A.S. and Kurths, J., 1997). One example of controlling the amount of noise in a system to obtain beneficial effects is that of simulated annealing, in which the temperature of the model (scale factor in the energy function or equivalently the inverse of the learning rate) slowly cools-down to help the optimization process to explore a large region until settling in a good minimum (Kirkpatrick et al, 1983). Another example of constructive effect in the training of neural networks is the injection of additive noise to the input. This has been shown to result in a regularizing effect which favors the network to settle in wider local minima of the energy landscape (Hochreiter, S. And Schmidhuber, J., 1995). Such flatter minima solutions provide robustness of the learned model under some variations occurring during testing. Similarly, multiplicative noise can also result in strong regularizing effects (Srivastava et al, 2014). In this case, the injection of multiplicative binary noise at input and higher layers of the model, such as it occurs in dropout methods, can result in beneficial effects and increase robustness at different scales by acting at all levels of the representation hierarchy.

In general, many connections have surfaced between the objectives and methods of statistical physics and those of large collections of learning neurons. Originally, statistical physics was devised to deal with systems made up of an incredible amount of atoms or units without the need to describe the dynamics of each of them but just their statistical behavior. A similar perspective and the formal mappings with statistical models add another view on how coalitions of tens of thousands of neurons self-organize to produce large-scale computations. Energy landscapes, simulated annealing, temperature, Boltzmann sampling, noise, dynamical systems, or phase transitions are just some of the concepts that have successfully transferred from physics communities to machine learning and more recently to deep learning to improve and explain how it works.

## 2.6. Computational Perspective

Considering that any given neural network encodes a well-defined function, one can also view the processing by different layers as a series of computational steps which combination produces a desired output (Goodfellow et al, 2016). A series of simple steps building on top of previous ones is also how we usually find more natural to code a certain function or give complex instructions to a computer, possibly reusing definitions and outputs of previous computational steps to define new ones. Thus, it is perhaps natural to imagine that while the expressibility of a neural network does not increase beyond a single hidden layer, the learning of a complex function might be easier when assisted by a number of intermediate layers or steps.

More generally, modern neural networks possess increasingly complex and heterogeneous architectures and are now better understood as a special case of computational graphs (Goodfellow et al, 2016). In this framework, nodes of the computational graph represent different variables (scalars or tensors). Edges between nodes in the graph indicate how some variables are obtained from a set of simple operations on other variables or nodes. The graph can be of very complicated topology and contain different classes of operations. Thankfully, the apparatus for the efficient computation of back-propagation can be generalized to these computational graphs, which allows for general recipes of how to formalize and train modern neural networks. From a design point of view, these complex graphs can now be made at the level of connecting several high-level modules each with a well defined function within the architecture (e.g. vision module, attention module,...).

From a computational point of view one can also ask how hard is to train neural networks. That is to ask how the computational resources needed to solve the optimization problem during training scale with the size of the problem. It is known that finding the absolute minimum in the energy or loss landscape of networks with at least one hidden layer is an NP-problem (Blum, A. and Rivest, R.L., 1989). However, under some approximations it has been shown that there is a strong relation between the index of a critical point (number of dimensions across which the point behaves like a local maximum) and its height in the loss function (Dauphin et al, 2014). In particular, local minima (index=0) tend to concentrate at energy levels almost as low as the lowest height possible attained by the absolute minima, explaining why in practice an absolutely optimal set of weights does not need to be found. The study of critical points in loss landscapes, such as minima and saddle points, has provided insights into the workings of numerical optimization methods in deep learning. However, the understanding of the full trajectory of the network during learning, including the effects of the weights initialization, the speed of convergence, or number of hidden layers, will require a more detailed description of the loss landscape than just its critical points.

The lens of computational complexity (Wigderson, 2019) can also guide us to better compare across deep learning algorithms in their efficiency at solving tasks. Recently, several deep learning algorithms have even been ascribed the term of human-level or superhuman for their comparison to human performance when solving the same task. But it is essential that we distil how much of their success can be attributed to the data (amount, richness of learning experiences), the algorithm (architecture, learning and decision making rules), the computational resources (amount of memory, time, bandwidth), and even hardware (physical ability to respond fast and accurately). While performance can measure the overall success at solving a task, computational complexity reminds us that a fair comparison between algorithms requires also to quantify the computational resources consumed by the algorithms (in machines and humans) and how these scale up. This strategy will require to go beyond simply reporting the final performance and computer time needed to solve a task, but a systematic study of how the performance varies when changing the complexity of the problem and the available resources. Only then we will better understand where the merits and disadvantages of each system lay.

## 2.7. Neuroscience Perspective

The language of deep learning is filled with terms such as neural network, attention, short-term memory, episodic memory, reinforcement learning and many others which have been borrowed from neuroscience and cognitive sciences. Historically, human and animal brains and cognition has been the main source of inspiration for connectionist models such as deep learning. Indeed, the basic scheme that information is processed by nodes receiving inputs from other nodes, and that learning proceeds by modifications of such interconnections, possibly routing the information differently as one learns, goes back to the times of Ramón y Cajal when the modern study of neuroscience and psychology flourished. However, it took many decades to amass solid knowledge for neuroscience to inspire the first formal models and implementations of primitive artificial neural networks. In the hands of McCulloch, Pitts, and Rosenblath the fundamental ideas of such network models took shape and still remain the backbone of deep learning.

Since then more detailed phenomena and concepts first observed in neuroscience have also been used in artificial neural network models. Activation functions, such as ReLU among others (Hanhloser et al, 2000), connectivity patterns such as feedback connections (Lotter et al, 2016), or different plasticity and learning rules described in biological networks (Kirkpatrick et al, 2017) have been continuously transferred to their artificial counterparts. However, recreating a physiologically plausible brain might not always be the only guide and it can be a misleading strategy if overengineered (Theil, 2015). Biological networks evolved under different pressures and real neurons have a range of functions and constraints related to their biological origin that might not be wise to mimic in artificial networks in silico. Artificial neurons do not have to worry about metabolism, homeostasis or the general task of being alive. Hence, at the implementation level we face the challenge to distinguish, within the vast but incomplete sea of biological data, which biological processes realize an important computational function for the task at hand, and which reflect other functions or are simply side-effects. Luckily the transfer of ideas can also occur at higher levels of analysis than the implementation level (Marr, D. and Poggio, T., 1976).

Indeed, psychology and cognitive science have provided us with a rich source of inspiration for computational modules without an accompanying knowledge on how these are actually implemented in brains. Working at the level of representational structures and computational processes these disciplines have identified many necessary or sufficient functions for solving different tasks without an attachment to any specific "hardware" or mechanistic implementation. Thus, constructs relying on computational and representational structures such as attention layers, saliency maps, or different types of memory have been incorporated as key components of deep learning models (Hassabis et al, 2017). Indeed, the design and understanding of complex deep learning architectures is often driven by intuitions on such functional and modular levels while abstracting from particular implementations.

Methods to analyze biological neural signals have also being transferred to study artificial networks. For example, the concept of receptive fields has been widely used in neuroscience to characterize the activation of neurons by sensory stimuli. This concept has also been used to explain and visualize the activations of artificial neurons in the presence of a complex and specific constellation of stimuli. Such neurons are often realized near the output layer of a classifier and can be understood by the formation of a hierarchy of increasingly complex receptive fields from the conjunction of simpler ones at previous levels (Gross, 2002). Moreover, the discoveries by neuroscientists of biological neurons with striking receptive fields such as grid cells (Moser et al, 2008), mirror neurons (Rizzolatti, G., and Craighero, L., 2004), or simply cells that respond to a high-level concept regardless of the input modality (Quiroga et al, 2005), continue to inspire new computational models (Banino et al, 2018).

More than with any other discipline, the bridge between AI and neuroscience has been a two-way road with advances in each discipline fueling new ideas in the other. However, a word of caveat might be due. Brains and artificial networks share some obvious similarities but they are systems of different complexity operating on different substrates which evolved under different pressures and possess different computational resources. Therefore, not all the state of the art results obtained by deep learning must launch a desperate search for analog mechanisms in the brain. In the same way, not all brain processes must be mimicked to reverse-engineer human intelligence. Nevertheless, the computational and representational similarities between deep learning algorithms and brains when they are subject to similar tasks and computational resources can be an excellent guide for better understanding how each system works at the algorithmic level, which can then constrain the space of possible mechanisms. Hopefully, crossing the bridge in both directions will bring us to narrow the gap in our understanding of what brains compute and create technology that improves our lives.

## 3. Discussion

Looking at a problem through multiple angles can stimulate novel ideas. Our discussion on each perspective and examples of intuitions have been necessarily shallow due to the multiple views that had to be covered. The deepness in this case should come from putting all these faces of deep learning together in the reader's mind and entertain their interrelations. We hope that such multiple views will help practitioners to get interested in different communities to complement their own views and intuitions on how deep learning works.

## 4. Conflict of Interest

*The authors declare that the research was conducted in the absence of any commercial or financial relationships that could be construed as a potential conflict of interest.*

## 5. Funding

RV thanks the financial support by the Estonian Research Council (project number PUT 1476), and the Estonian Centre of Excellence in IT (EXCITE) (project number TK148).

## 6. Acknowledgments

RV thanks all members of the group of computational neuroscience (Jaan Aru, Tambet Matiisen, Ardi Tampuu, Ilya Kuzovkin, Daniel Majoral, Aqeel Labash, Oriol Andreu, Roman Ring, Kristjan Korjus, Abdullah Makkeh) for enlightening discussions and patient explanations over the years on many of the concepts covered in this article.